\pdfoutput=1
\documentclass[11pt]{article}

\usepackage[final]{acl}
\usepackage{float}
\usepackage{times}
\usepackage{latexsym}
\usepackage{booktabs}
\usepackage{caption}
\usepackage{makecell,enumitem}

\usepackage{bm}
\usepackage{multirow}
\usepackage{listings}
\usepackage{blindtext}
\usepackage{wrapfig}
\usepackage{subcaption} % Add this to your preamble if it's not already included

\usepackage{amsmath}
\usepackage{colortbl}
\usepackage{algorithm}
\usepackage{algpseudocode}
\definecolor{mygray}{gray}{.91}
\usepackage[utf8]{inputenc} % allow utf-8 input
\usepackage[T1]{fontenc}    % use 8-bit T1 fonts
\usepackage{hyperref}       % hyperlinks
\usepackage{url}            % simple URL typesetting
\usepackage{booktabs}       % professional-quality tables
\usepackage{amsfonts}       % blackboard math symbols
\usepackage{nicefrac}       % compact symbols for 1/2, etc.
\usepackage{xcolor}         % colors

\lstdefinestyle{python}{
    language=Python,
    basicstyle=\ttfamily\small,
    keywordstyle=\bfseries\color{blue},
    commentstyle=\itshape\color{teal}, 
    stringstyle=\color{orange},
    showstringspaces=false,
    numbers=left,
    numberstyle=\tiny,
    stepnumber=1,
    numbersep=3pt,
    breaklines=true,
    tabsize=4,
    columns=flexible,
    morekeywords={self},
}
\usepackage[T1]{fontenc}
\usepackage[utf8]{inputenc}

\usepackage{microtype}

\usepackage{inconsolata}

\usepackage{graphicx}

\title{Wanda++: Pruning Large Language Models via Regional Gradients}

\author{
  \vspace{-5mm}\\\footnotesize \textbf{Yifan Yang\textsuperscript{$\diamondsuit$,\textdagger,*,\textdaggerdbl}~~~Kai Zhen\textsuperscript{$\clubsuit$,\textdagger,\textdaggerdbl}~~~Bhavana Ganesh\textsuperscript{$\clubsuit$}~~~Aram Galstyan\textsuperscript{$\clubsuit$}~~~Goeric Huybrechts\textsuperscript{$\clubsuit$}}\\
  \small \textbf{Markus Müller\textsuperscript{$\clubsuit$}~~~Jonas M. Kübler\textsuperscript{$\clubsuit$}~~~Rupak Vignesh Swaminathan\textsuperscript{$\clubsuit$}~~~Athanasios Mouchtaris\textsuperscript{$\clubsuit$}}\\
  \small \textbf{Sravan Babu Bodapati\textsuperscript{$\clubsuit$}~~~Nathan Susanj\textsuperscript{$\clubsuit$}~~~Zheng Zhang\textsuperscript{$\diamondsuit$}~~~Jack FitzGerald\textsuperscript{$\clubsuit$}~~~Abhishek Kumar\textsuperscript{$\clubsuit$}}\\[.5mm]
  \small \textsuperscript{$\diamondsuit$} University of California, Santa Barbara  \quad \small \textsuperscript{$\clubsuit$} Amazon AGI \\
  \small \textsuperscript{\textdagger}Equal contributions \textsuperscript{*} Work done at Amazon \\
  \small \textsuperscript{\textdaggerdbl} Corresponding authors: \texttt{yifanyang@cs.ucsb.edu, kaizhen@amazon.com}\\
  \small Project Page: \url{https://yifan-yang.net/wandapp.github.io/}
}

\begin{document}
\maketitle
\begin{abstract}

Large Language Models (LLMs) pruning seeks to remove unimportant weights for inference speedup with minimal accuracy impact. However, existing methods often suffer from accuracy degradation without full-model sparsity-aware fine-tuning. This paper presents Wanda++, a novel pruning framework that outperforms the state-of-the-art methods by utilizing decoder-block-level \textbf{regional} gradients. Specifically, Wanda++ improves the pruning score with regional gradients for the first time and proposes an efficient regional optimization method to minimize pruning-induced output discrepancies between the dense and sparse decoder output. Notably, Wanda++ improves perplexity by up to 32\% over Wanda in the language modeling task and generalizes effectively to downstream tasks. Moreover, despite updating weights with regional optimization, Wanda++ remains orthogonal to sparsity-aware fine-tuning, further reducing perplexity with LoRA in great extend. Our approach is lightweight, pruning a 7B LLaMA model in under 10 minutes on a single H100 GPU.

% As Large Language Models (LLMs) continue to grow in size, retraining-free pruning methods are increasingly used for inference acceleration for their memory and data efficiency. Yet, these methods often notably degrade performance by ignoring informative gradients from non-linear layers. However, obtaining such information through full model backpropagation, as in previous approaches, has become increasingly impractical. In this paper, we propose a new efficient pruning framework Wanda++ to achieve state-of-the-art pruning performance. Specifically, we consider Wanda \citep{sun2023simple} as a representative pruning baseline, identify flaws in its pruning criterion, and introduce a new criterion that leverages decoder-level regional gradients.Also, we further consider a regional optimization process to quickly update the model weights with only the regional gradient information. We evaluate Wanda++’s next-token prediction capabilities on LLaMA-1 and OpenLLaMA model families, and consistently observed >20\% relative perplexity improvement over Wanda method.  Further zero-shot downstream tasks evaluation shows the strong generalization ability of the proposed Wanda++ framework, maintaining comparable accuracies in \textsc{MMLU}, \textsc{QA}, and \textsc{NLI} tasks. Our method achieves the optimal performance with extreme data efficiency and can perform state-of-the-art pruning on OpenLLaMA-7B within 1 hour on a single NVIDIA H100 GPU.
\end{abstract}

\section{Introduction}
%Large Language Models (LLMs) have demonstrated remarkable task generalization capabilities across various applications \citep{devlin2018bert,touvron2023LLaMA}. However, the rapidly increasing size of these models leads to significant memory and computational costs for inference. For instance, the OpenLLaMA model with 70 billion parameters, requires at least 140GB of GPU memory in FP16 format and needs at least four H100 GPUs, each with 80GB of memory, for inference. To accelerate inference and reduce memory costs, various model compression approaches have been explored for LLMs, including weight decomposition \citep{hsu2022language, yang2024loretta}, quantization \citep{lin2024awq}, and pruning \citep{sun2023simple, frantar2023sparsegpt}.

The growing size of Large Language Models (LLMs) improves performance \citep{touvron2023LLaMA, Intelligence2024, Intelligence2025} at the cost of memory consumption and inference latency. For example, loading the weights of LLaMA-2 70B requires 140 GB of GPU memory in FP16 format. Hosting a such model even with a batch size of 1 and 512 input tokens needs at least four A100-40GB GPUs, with the time to first token (TTFT) exceeding 100 milliseconds \citep{Agarwal2023}. To address these challenges, various model compression approaches, including weight decomposition \citep{hsu2022language, yang2024loretta, ghiasvand2024communication}, quantization \citep{lin2024awq, zhou2025quzo}, and pruning \citep{sun2023simple, frantar2023sparsegpt, zhang2024sparse}, have been explored.
%Among the various model compression methods, pruning—specifically, shrinks the network size by removing weights elements or entire modules from the model—has garnered significant attention prior to the era of LLMs \citep{han2015learning, frankle2018lottery}. However, traditional pruning techniques struggle to adapt to recent gigantic LLMs since they often necessitate complete model retraining, which is infeasible with limited computational resources. To address this challenge, a recent trend in LLMs pruning involves post-training pruning without retraining, as seen in methods like SparseGPT \citep{frantar2023sparsegpt} and Wanda \citep{sun2023simple}. Despite this innovation, these methods have yet to achieve satisfactory performance in their pruning results.\\~\\
Similar to LLM quantization, many recent LLM pruning methods have shifted from in-training approaches \citep{han2015learning, frankle2018lottery} to post-training methods, as seen in SparseGPT \citep{frantar2023sparsegpt} and Wanda \citep{sun2023simple}. However, unlike post-training quantization like AWQ \citep{lin2024awq}, which compresses model weights almost losslessly by 4$\times$ (from 16-bit to 4-bit), these pruning methods have yet to achieve comparable levels of performance.
\begin{figure}[t]
    \centering
\includegraphics[width=0.47\textwidth]{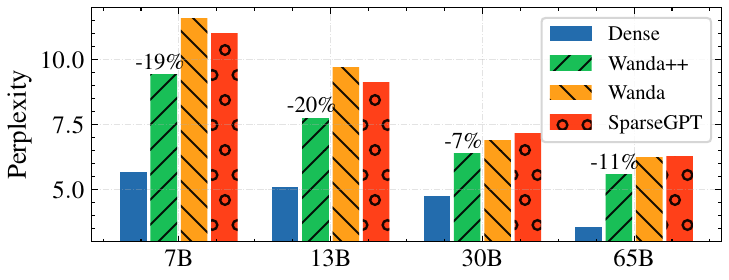}\vspace{-5pt}
    \caption{Wanda++ mitigates 2:4 pruning-induced degradation more effectively, with relative perplexity improvement over Wanda shown on Wikitext using LLaMA-1 models across four different sizes.}
    \label{fig:first}
    \vspace{-10pt}
\end{figure}

To mitigate the performance degradation, GBLM and Pruner-Zero \citep{das2023beyond, dong2024prunerzero} propose improved pruning criteria that enhance the layer-wise Wanda score by incorporating gradient information obtained through full-model backpropagation (BP). These studies highlight the importance of gradient information, demonstrating that it provides valuable insights even though Wanda assumes that the gradients of a fully trained network are small and contribute little when higher-order terms are considered. Meanwhile, other approaches focus on recovering model performance through full-model sparsity-aware tuning \cite{sun2023simple} or distillation \citep{liang2023homodistil}. However, all these methods, which heavily depend on full-model loss, suffer from impractical memory requirements and excessive pruning time due to the high computational cost of full-model backpropagation. This raises the question: 

\textit{Is there a way to effectively involve gradient information while still in a lightweight manner?}%\\~\\

In this paper, we propose Wanda++ pruning framework to leverage gradients at the ``decoder-block" level, termed regional gradients, which shows significant improvement compared to Wanda \cite{sun2023simple} without sacrificing the pruning efficiency. Compared with traditional full model BP, the regional gradient can be obtained by only loading and computing gradients per single decoder block, which largely reduce the GPU hours required. Compare with previous gradient-free method like Wanda, the regional gradient can improve the pruning performance by accessing the gradient information. Wanda++ prune each decoder block by iteratively prune the model based on a Regional Gradient Score (RGS) and slightly update the weights with our proposed Regional Optimizer (RO).

To effectively obtain the regional gradient in the RGS score, we compute the regional gradient through a backward process on a regional loss, which is defined as the $\ell_2$ norm of each decoder’s output hidden states. To involve the regional gradient into the pruning score, we follow the design of GBLM score \cite{das2023beyond}. Note that the regional gradient is only computed once during the iterative pruning process of each decoder block and reused for all RO iterations to further reduce the computation. As a result, the regional gradient may become less accurate after weights being updated during the RO process. The design of GBLM score provide an effective way to reflect changes by fetch the input per in-block layer and blend it into the RGS along with regional gradients.  

For the RO in Wanda++, we construct a simple loss function between the outputs of the dense and pruned decoder blocks to update the model weights within each block. The concept of regional optimization traces back to previous local optimization approaches in convolutional neural networks (CNNs), which aimed to optimize models by focusing on individual convolutional layers \citep{wang2021revisiting}. Differently, RO in Wanda++ extends this idea for mitigating pruning-induced loss on a small portion of calibration data at each LLM decoder-block, instead of optimizing the cross-entropy loss for classification tasks.

Our proposed framework achieves up to a 32\% reduction in WikiText perplexity at 2:4 sparsity, compared to the Wanda method under the same experimental setup, as shown in Figure. \ref{fig:first}. Our contributions can be summarized as follows:
\begin{itemize}[leftmargin=*]
    \item We propose Wanda++, a lightweight yet effective framework that prunes LLMs using regional gradients, achieving performance improvements without requiring access to full model gradients.
    \item Wanda++ effectively lowers the pruning-induced degradation in a non-incremental way and generalize well in zero-shot downstream tasks.
    \item The RO method in Wanda++ is orthogonal to previous full-model sparsity-aware fine-tuning methods and has been shown to achieve a similar perplexity improvement as Wanda after applying LoRA fine-tuning.
    %Comprehensive studies are conducted to showcase the effectiveness of the proposed method, including the efficiency, and performance under higher sparsity, quantization cases.
\end{itemize}

% \begin{wrapfigure}{r}{0.25\textwidth} %this figure will be at the right
%     \centering
%     \caption{Wikitext perplexity comparison on LLaMA-2-7B model with 2:4 and 4: sparsity.}
%     \label{fig:first}
%     \includegraphics[width=0.25\textwidth]{mesh}
% \end{wrapfigure}

% To show the effectiveness of the Wanda++ method, we conduct comprehensive experiments on the widely adopted Llama-2 model families. Our results shows that the proposed Wanda++ method achieves remarkable improvement than the original Wanda method, with same number of GPU required for 7B model and only one additional GPU required for 70B model. Additional results on zero-shot downstream tasks, pruning memory/time comparison also shows the effectiveness and efficiency for considering the regional gradient information to improve the pruning performance.
% \section{Related works}
% \label{gen_inst}
% \section{$\textbf{Wanda}^{{\color{red}{++}}}_{{\color{blue}{++}}}$}
% \label{headings}
\section{Related Work}
\textbf{Network Pruning:} The concept of pruning neural networks has been explored for decades, beginning with foundational works such as \citep{lecun1989optimal, hassibi1993optimal}. In addition to widely studied unstructured pruning approaches, structured pruning methods, which remove entire subnets of a network or rows/columns within weight matrices, are more easily supported on hardware for inference speedup \citep{liu2017learning, shen2022structural}. Research in this area has focused on analyzing input/output activation statistics to identify the most suitable neurons for pruning \citep{bai2021explainable, molchanov2022pruning}. However, even at 50\% sparsity, structured pruning often results in non-trivial performance degradation \citep{ashkboos2024slicegpt}.
Beyond these methods, semi-structured pruning \citep{pool2021accelerating, fang2022algorithm, fang2023efficient}, such as 2:4 sparsity, shows greater resilience at 50\% sparsity and can effectively reduce runtime latency with throughput improvement on NVIDIA's recent hardware. Therefore, this work primarily focuses on unstructured and semi-structured patterns, although the proposed Wanda++, RO in particular, is sparsity pattern agnostic.

\noindent\textbf{LLM Pruning:} As LLMs continue to grow in size, scaling traditional pruning methods to accommodate them presents significant challenges. Traditional pruning methods, which typically require full model retraining, demand substantial computational resources, making them impractical in the era of LLMs. A notable trend in LLM pruning is the adoption of post-training pruning methods \citep{frantar2023sparsegpt, sun2023simple, das2023beyond, zhang2025mazo}, which develop specific pruning scores to determine the importance of different weight elements. Additionally, SliceGPT \citep{ashkboos2024slicegpt} focuses on structured pruning by slicing rows or columns of weight matrices based on input eigenvectors and eigenvalues.

\noindent\textbf{Pruning with Gradient Information:} The use of gradient information during the training process has been extensively studied in two ways.  The first set of methods directly incorporates gradient information to refine their pruning scores at a finer granularity. Recent developments, such as GBLM and Pruner-Zero \citep{das2023beyond, dong2024prunerzero} use gradients obtained through the full model backward with respect to cross entropy loss to achieve better performance, though they are time and memory infeasible for large-scale model.

The second set of methods focuses more on incorporating gradient information during sparsity-aware distillation or fine-tuning. For example, \cite{liang2023homodistil} explores a task-agnostic distillation approach combined with iterative pruning to achieve strong performance on natural language understanding tasks. Additionally, full-model or LoRA fine-tuning has been considered as an auxiliary step in various prior pruning works that emphasize single-shot performance to further reduce the pruning degradation, such as Wanda \cite{sun2023simple} and SliceGPT \cite{ashkboos2024slicegpt}.
\section{Prior Solutions}\label{sec:wanda}
Existing solvers for post-training LLM pruning often start with magnitude pruning \citep{han2015learning}, which removes individual weights based on their magnitudes. This approach has been widely adopted as a baseline method in both vision \citep{han2015deep} and language processing \citep{gale2019state}. Magnitude pruning is built on the assumption that neurons with larger weight elements contribute higher gradients when input features have similar magnitudes. However, this assumption does not always hold true for LLMs, where input features can differ significantly in scale, as observed in \citep{dettmers2022gptint}.

To address this challenge, Sun et al. propose the Wanda method \citep{sun2023simple}, which introduces a pruning criterion that multiplies the weight magnitude by the input activation on an element-wise basis. Specifically, given a linear layer with a weight matrix $\bm{W}\in\mathbb{R}^{d_{out} \times d_{in}}$ and input $\bm{X}\in\mathbb{R}^{(B \times L) \times d_{in}}$ assuming the batch size is $B$, sequence length is $L$, the Wanda pruning score for the weight element connect input $j$ and output $i$ is defined as:
\begin{align}
    \bm{S}_{ij} = |\bm{W}_{ij}|\cdot\|\bm{X}_j\|_2,
\end{align}
%where $\|\bm{X}_j\|_2$ represent the L2 norm of the input $\bm{X}$. 
where $\|\bm{X}_j\|_2$ represent the L2 norm of the $j$-th input channel $\bm{X}_j$. The Wanda score is further improved by \cite{das2023beyond}, which shows the first order gradient is still crucial even though Wanda score consider the higher-order terms and gives the GBLM score:
\begin{align}%\label{eq:rgs}
    \bm{S}_{ij} = (\alpha \bm{G}_{ij} + \|\bm{X}_j\|_2)\cdot |\bm{W}_{ij}|,
\end{align}
where $\bm{G}$ is full model gradient with respect to the cross-entropy loss between the prediction and labels. Next, we introduce how to involve the regional gradient into the pruning score and perform a effective and lightweight regional optimization.

% However, we argue that this formulation has two problems: 
% \begin{itemize}[leftmargin=*]
%     \item Wanda assumes that the model is linear w.r.t. its parameters $\bm{W}$ (as we shown in Sec. \ref{sec:theory}) .
%     \item $\|\bm{X}_j\|_2$ grows with the number of examples which makes the criterion sensitive to the amount of data used in a non-trivial way. 
% \end{itemize}Next, we explain how the Wanda score is derived using a first-order Taylor expansion under an over-optimistic linearity assumption. We then introduce Wanda++, which accounts for non-linearity by incorporating regional gradients during both the pruning and optimization stages.
\begin{figure*}[t]
    \centering
    \includegraphics[width=0.9\textwidth]{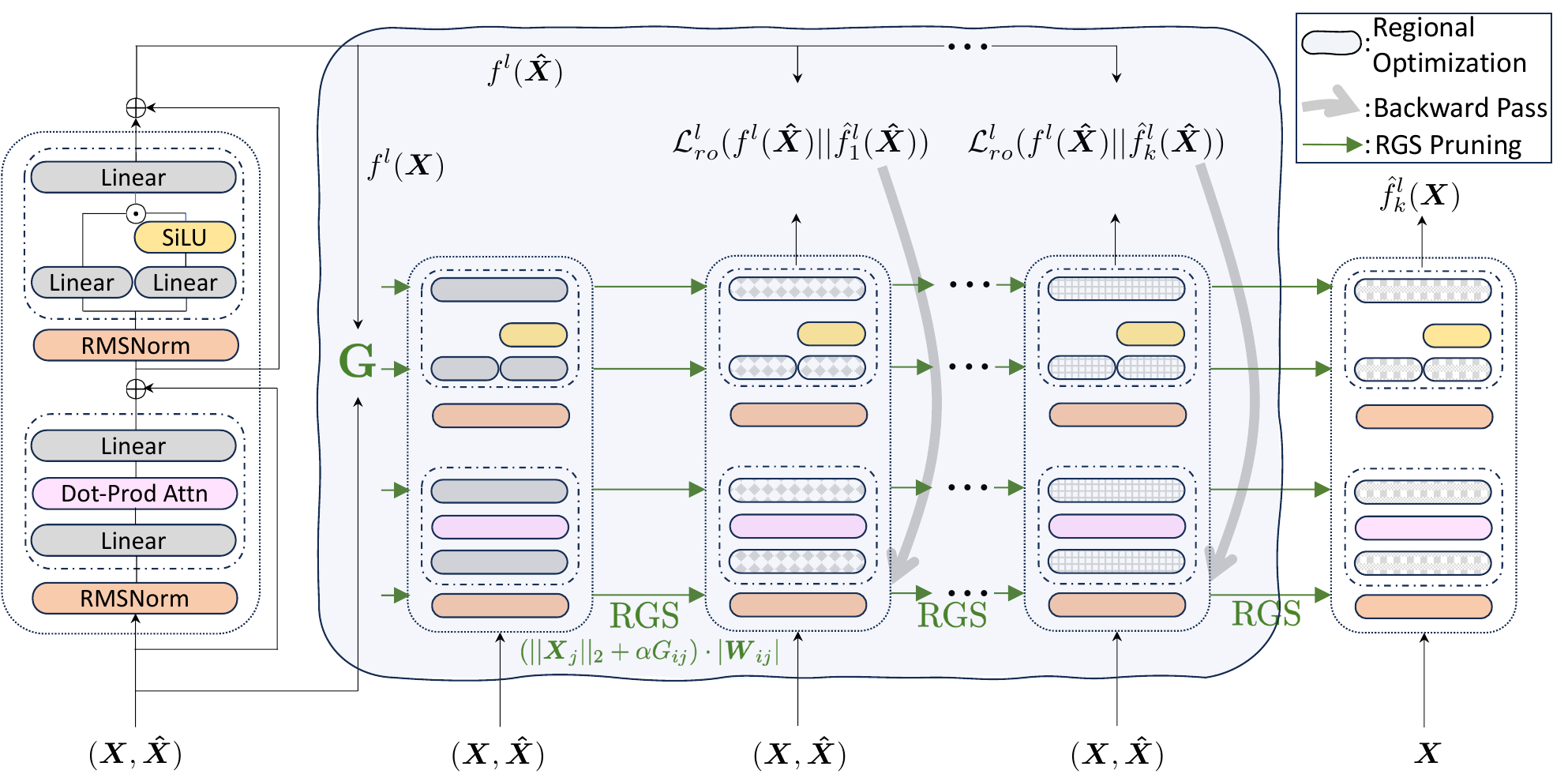}\vspace{-1pt}
    \caption{
    %\textbf{(TBD: need some figure including both RGS and RO process) }Illustration of the RGS criterion. The regional gradient $G_W$ of the decoder output L2 norm is obtained by some local BP through each decoder block.
    Illustration of Wanda++, which leverages regional gradients in two ways: first, we estimate the block-level gradient $\mathbf{G}$, used in Regional-Gradient Scoring (RGS) for layer-wise pruning in each block. Next, we calculate the pruning-induced loss $\mathcal{L}{ro}$ for lightweight Regional Optimization (RO). RO can be iterated multiple times to infer better pruning masks for each decoder block.
    }
    \label{fig:methods}
    \vspace{-10pt}
\end{figure*}
\section{The Wanda++ Framework}
%As noted in the introduction, traditional memory-efficient pruning methods often lead to performance degradation in pruned models due to the linearity assumption of the model function. 
In this section, we use the Wanda method to highlight the drawbacks of the linearity assumption in previous layer-wise post-training pruning approaches. We then introduce our Wanda++ framework, which efficiently reduces pruning degradation with a two-stage process for each decoder. This involves layer-wise pruning within blocks based on a new pruning criterion called the Regional Gradient Score (RGS) and Regional Optimization (RO), as shown in Figure \ref{fig:methods}.
%In this section, we use the Wanda method as an example to analyze the drawbacks of the linearity assumption of the model function from previous layer-wise post-training pruning approaches. We then introduce our Wanda++ framework, which efficiently mitigates pruning degradation through an iterative two-stage process for each decoder, which involves in-block layer-wise pruning based on the novel pruning criterion of Regional Gradient Score (RGS)  and Regional Optimization (RO), as shown in Figure. \ref{fig:methods}. 
Wanda++ is efficient as it operates at each decoder block where its RO requires only 5 iterations.

\subsection{Regional Gradient Score}
As the first stage of Wanda++, we obtain the Regional Gradient Score (RGS) for in-block layer-wise pruning. We start with constructing an RGS loss function for obtaining the gradient of each weight matrix. Given a model with $L$ decoder blocks, we represent the set of input hidden states for the $l$-th decoder block specifically as $\mathcal{X}^l=\{\bm{X}_1^l,\cdots, \bm{X}_N^l\}$ and define the decoder block function as $f^l(\bm{X}^l_n)$ with input $\bm{X}^l_n\in \mathcal{X}^l$. The RGS loss for $l$-th block is defined as $\mathcal{L}_{RGS}^l(\bm{X}^l_n) = \|f^l(\bm{X}^l_n)\|_2$. By performing a single BP through a certain decoder block regarding the regional loss $\mathcal{L}^l$, we can efficiently obtain the stochastic estimated gradient magnitude for each weight matrix by taking the absolute value for the estimated gradient $\nabla_{\bm{W}_{ij}} \mathcal{L}_{RGS}^l(\bm{X}^l_n)$.

By performing the element-wise multiplication between the weight and estimated gradient magnitude, we have the initial regional gradient-based score for each weight element $W_{ij}$:
\begin{align}\label{eq:score}
    \bm{S}_{ij} = (\sqrt{\frac{\sum^N_{n=1} \nabla\mathcal{L}_{RGS}^l(\bm{X}^l_n)^2}{N} })_{ij}\cdot |\bm{W}_{ij}|,
\end{align}
where $N$ represents the total number of input samples used for pruning. For simplicity, we record the score in Eq. (\ref{eq:score}) as $\bm{S}_{ij} = \bm{G}_{ij}\cdot|\bm{W}_{ij}|$.

To limit the overhead of computing the pruning criterion, we estimate the regional gradient (Eq. \ref{eq:rgs}) only once per decoder block. This can be sub-optimal because pruning one layer in the $l$-th block affects the regional gradients of other layers. To address this, we blend in the layer-wise Wanda score, which tracks how pruning in one layer impacts the input of the next. Our RGS criterion is summarized as follows:
\begin{align}%\label{eq:rgs}
    \bm{S}_{ij} = (\alpha \bm{G}_{ij} + \|\bm{X}_j\|_2)\cdot |\bm{W}_{ij}|,
    \label{eq:rgs}
\end{align}
where the scaling factor $\alpha$ is a constant to balance the magnitude of the gradient and input activation terms. For simplicity, we adopt the same scaling value of 100 as in \citep{das2023beyond} although it can be model specific (see the ablation study on $\alpha$ in Appendix \ref{app:alpha}).

\subsection{Regional Optimization}
The second stage for our Wanda++ framework is the Regional Optimization (RO). During this process, we slightly update the model weights within each decoder block to minimize the difference between the output from dense and pruned decoding blocks. Specifically, for the $l$-th decoder block, the output of the dense output can be represented as $f^l(\bm{X}^l_n)$ and the pruned output at $k$-th round is defined as $\hat{f}^{l,k}(\bm{X}^l_n)$, respectively. To further reduce the time of the RO process, we randomly select $M$ inputs from the inputs set $\mathcal{X}^l$ of each decoder block to construct an RO inputs set $\hat{\mathcal{X}}^l=\{\hat{\bm{X}}^l_1, \cdots, \hat{\bm{X}}^l_M\}$, without replacement. Then, the RO loss with input $\hat{\bm{X}}^l_m \in \hat{\mathcal{X}}^l$ for the $l$-th decoder in the $k$-th round can be defined as an MSE loss between the dense and pruned outputs, which gives:
\begin{align}
    \mathcal{L}^{l,k}_{ro}(\hat{\bm{X}}^l_m) = (f^l(\hat{\bm{X}}^l_m) - \hat{f}_k^l(\hat{\bm{X}}^l_m))^2.
\end{align}
For each RO sample $\hat{\bm{X}}^l_m$, we perform a forward pass within the decoder block to compute the RO loss, followed by backpropagation and a weight update. This process takes place after the pruning stage in each iteration of our Wanda++ framework. Typically, we randomly select 32 RO inputs from the 128 inputs used in the pruning stage at the start of each RO iteration. RMSprop optimizer \citep{ruder2016overview} is used with the learning rate of 3e-7.
\subsection{Algorithms}
We summarize the Wanda++ algorithm flow in Alg. \ref{alg:main}, as also shown in Figure \ref{fig:methods}. For each decoder block, we perform layer-wise RGS pruning (step 5) followed by a round of the RO process (steps 6-8) for $K$ iterations. An additional RGS pruning (step 11) is required to restore sparsity after RO. To detail how RGS is computed in steps 5 and 11, we provide PyTorch pseudo-code in Appendix \ref{app:algotihms}.
% \end{document}
% \textbf{Additional Memory Required by Considering Gradient:}
 
% \textbf{Fuse the layernorm}

% \textbf{Scaling factor}
% \textbf{Regional Optimization:} Although the newly introduced \textit{WandApp} method has achieved satisfactory improvements over the previous Wanda method, we are investigating whether further reductions in pruning degradation can be achieved by employing a Regional Optimization (RO) strategy within each decoder block. To facilitate the RO method, we define a regional loss between the outputs of the dense and pruned models for each decoder block $l$, expressed as:
% \begin{align}
%     \mathcal{L}^l = \|\text{(expression for regional loss)}\|_2,
% \end{align}
% To enhance the effectiveness of the RO method, we have designed an iterative pruning and optimization framework, as illustrated in Figure. \ref{fig:methods}. To further reduce the computation time of the RO method, we observe that maintaining the same regional gradient information throughout the entire RO process for each layer does not significantly impact performance. Therefore, we initially perform regional backpropagation to obtain a gradient dictionary for each weight matrix in the decoder module. Subsequently, we iteratively prune and optimize the weights based on the new weight magnitudes at each iteration.
\begin{algorithm*}[ht]
\caption{Pruning framework of Wanda++}
\label{alg:main}
\begin{algorithmic}[1]
\Require $\{\mathcal{X}^l\}^{l\in[1,L]}$ \textcolor{blue}{\Comment{Inputs set for each decoder block}}
% \Require $\{\bm{W}^l_q, \bm{W}^l_k, \bm{W}^l_v, \bm{W}^l_o, \bm{W}^l_{up}, \bm{W}^l_{gate}, \bm{W}^l_{down}\}^{l\in[1,L]}$ \textcolor{blue}{\Comment{List of weights in the $l$-th decoder block}}
\Require Scaling factor $\alpha$

\For{$\ell = 1 \ldots L$}
\State Calculating the RGS loss $\mathcal{L}_{RGS}^l$ with $\mathcal{X}^l$, backward, and record gradient $G$
\For{$k = 1 \ldots K$}
\State Selecting RO samples $\hat{\mathcal{X}}^l$ from $\mathcal{X}^l$
    \State Calculating and pruning with RGS by Eq. (\ref{eq:rgs}) \textcolor{blue}{\Comment{Stage 1: Pruning}}
    \For{$\hat{\bm{X}}^l_m\in\hat{\mathcal{X}}^l$} \textcolor{blue}{\Comment{Stage 2: Regional Optimization}}
  \State Calculating the RO loss $ \mathcal{L}^{l,k}_{ro}(\hat{\bm{X}}^l_m)$, backward, and update weights
\EndFor
\EndFor
\State Calculating the RGS loss $\mathcal{L}_{RGS}^l$ with $\mathcal{X}^l$, backward, and record gradient $G$
    \State Calculating and pruning with RGS by Eq. (\ref{eq:rgs}) 
\EndFor
\State \Return Pruned model
\end{algorithmic}
\vspace{-3pt}
\end{algorithm*}
\section{Experiment}
%\zz{In the result section, you mention two proposed methods: Wanda ++ RO and Wanda ++ RGS. But these two methods are NOT described separately as two different methods in the previous sections. Instead, they are described as two stages of one method. }
We evaluate the proposed pruning method based on four criteria. The first is perplexity in next-token prediction, which is the main objective of LLM pre-training. A lower perplexity is often a strong indicator of superior performance in language understanding tasks. The second criterion is the performance on zero-shot/few-shot NLP tasks, such as text classification, question answering, and text generation. This is to ensure that the gradients involved in regional optimization do not lead to overfitting for downstream tasks. The third criterion is pruning time and memory consumption. Finally, we examine the actual latency reduction under both FP16 and FP8 quantization settings for various batch sizes and input/output lengths.
\subsection{Experimental Setup}
Regarding the model, we consider OpenLLaMA (3B/7B/70B), LLaMA-1 (7B/13B/30B/65B) and LLaMA-3.1 (8B). 
%Due to licensing restrictions for LLaMA-2 and LLaMA-3.1, we do not report results from these models, although the proposed method is generic and applicable to them as well. 
We follow the same settings as in \citep{sun2023simple} to examine pruning performance on unstructured sparsity, 2:4 sparsity, and 4:8 sparsity. By default, we randomly select 128 samples from the C4 training data for regional optimization and evaluate perplexity on both the C4 validation and Wikitext test datasets. For zero-shot evaluation, we use the Harness evaluation toolkit \citep{eval-harness}. All experiments are conducted on the NVIDIA H100 GPU, where a single GPU is sufficient for models with 13B or fewer parameters and 4 GPUs are used for 65B and 70B models. Latency and model size are measured via TensorRT-LLM. 
Regarding the proposed pruning methods, by default, Wanda++ enables both RGS and RO; Wanda++ RGS excludes RO, using only RGS for pruning, while Wanda++ RO applies the Wanda score \citep{sun2023simple} for pruning and weight updates within each decoder block.
%\section{Methods}
%\begin{figure}[t]
%    \centering
%    \begin{subfigure}[t]{0.35\textwidth}
%        \centering
%        \includegraphics[width=\textwidth]{Fig/ro_curve_openllama_3b_c4.pdf}
%        %\caption{Illustration of our Wanda++ method. The regional gradient $G_W$ of the decoder output L2 norm is obtained by some local backpropagation through each decoder block. Then, the pruning score is computed by element-wise production between the weight and gradient magnitude, where the scaling factor $\alpha$ is used to control the relative importance between the gradient and weight magnitude.}
%        \label{fig:ro_curve_top}
%    \end{subfigure}
%    
%    \vspace{-0.1cm} % Adjust vertical spacing between subfigures if needed
%    
%    \begin{subfigure}[b]{0.35\textwidth}
%        \centering
%        \includegraphics[width=\textwidth]{Fig/ro_curve_openllama_3b_wikitext.pdf} % Replace with your second figure file
%        %\caption{Another illustration demonstrating the method with different parameters.}
%        \label{fig:ro_curve_bottom}
%    \end{subfigure}
%    
%    \caption{Illustration of our Wanda++ method. The regional gradient $G_W$ of the decoder output L2 norm is obtained by some local backprogation through each decoder block. Then, the pruning score is computed by element-wise production between the weight and gradient magnitude, where the scaling factor $\alpha$ is used to control the relative importance between the gradient and weight magnitude.}
%    \label{fig:ro_curve_combined}
%\end{figure}

\begin{table*}
\centering
\footnotesize
\setlength\tabcolsep{1pt}

\resizebox{1\textwidth}{!}{%
\begin{tabular}{c|c|cccccccc}
%\hline
\toprule
\multirow{2}{*}{\textbf{Method}} & \multirow{2}{*}{\textbf{Sparsity}} & \multicolumn{4}{c}{\textbf{LLaMA-1}} & \multicolumn{3}{c}{\textbf{OpenLLaMA}} & \textbf{LLaMA-3.1} \\
\cmidrule(lr){3-10} %\cmidrule(lr){7-}
 & & 7B & 13B & 30B & 65B & 3B & 7B & 70B & 8B \\
\midrule
Baseline   & -                    & 5.68   & 5.09 & 4.77 & 3.56 & 7.27    & 6.49    & 4.30   & 6.39 \\ \midrule
SparseGPT* & \multirow{6}{*}{0.5} & 7.22   & 6.21   & 5.31   & 4.57   & 10.41      & 8.57     & -  & - \\
Wanda*      &                       & 7.26   & 6.15   & 5.24   & 4.57   & 12.37      & 9.15    & 5.25  & 9.99 \\
GBLM &     &  7.15      & 6.11 &  5.18 &  -     & 10.75      &  8.49       & -    &           9.90
\\ 
\textit{Wanda++  RO} &             & 7.07   & 6.08   & 5.12   & 4.43   & 9.86      & 8.27   & 5.14   & 9.34%3432426452637 
\\ 
\textit{Wanda++  RGS}    &             & 7.18   & 6.12   & 5.15   & 4.48   & 10.78      & 8.50    & 5.19   & 9.92 \\
\textit{Wanda++} &          & {7.02 (-3\%)} & {6.00 (-2\%)} & {5.10 (-3\%)} & {4.43 (-3\%)} & \textbf{9.25 (-25\%)} & \textbf{7.82 (-15\%)}  & {5.11 (-3\%)} & \textbf{9.22 (-7\%)} \\ 
\midrule
SparseGPT* & \multirow{6}{*}{2:4} & 11.00  & 9.11   & 7.16   & 6.28   & 15.91      & 11.62    & -   & - \\
Wanda*      &                       & 11.53  & 9.58   & 6.90   & 6.25   & 28.04  & 15.35 & 6.47   & 24.83 \\
GBLM &    &  11.33        & 9.16 &6.87&    -   & 24.75  & 13.19 &-   & 24.34      \\ 
\textit{Wanda++  RO} &             & 10.78  & 7.89      &  6.51      & 5.86      & 19.41  & 11.69 & 6.37   & 19.43 \\ 
\textit{Wanda++  RGS}    &             & 11.46  & 9.44   & 6.93   & 6.23   & 24.77  & 13.27 & 6.40   & 24.54 \\
\textit{Wanda++} &          & \textbf{9.43 (-19\%)} & \textbf{7.75 (-20\%)} & \textbf{6.39 (-7\%)} & \textbf{5.59 (-11\%)} & \textbf{19.03 (-32\%)} & \textbf{11.30 (-26\%)} & {6.35 (-2\%)} & \textbf{18.32 (-26\%)} \\ 
\midrule
SparseGPT* & \multirow{6}{*}{4:8} & 8.61   & 7.40   & 6.17   & 5.38   & 12.20      & 9.79    & -   & - \\
Wanda*      &                       & 8.57   & 7.40   & 5.97   & 5.30   & 16.83  & 11.38 & 5.73     & 14.63 \\
GBLM &    & 8.48         & 7.26 &  5.89 &   -    & 14.86  & 10.38 & -   & 14.29 \\ 
\textit{Wanda++  RO} &             & 8.34   & 7.18      & 5.73      & 5.11      & 13.10  & 9.52  & 5.67   & 12.88 \\ 
\textit{Wanda++  RGS}    &             & 8.58   & 7.33   & 5.90   & 5.17   & 14.92  & 10.42 & 5.70   & 14.32 \\
\textit{Wanda++} &          & \textbf{7.88 (-8\%)} & \textbf{6.75 (-9\%)} & \textbf{5.65 (-5\%)} & {5.07 (-4\%)} & \textbf{12.54 (-25\%)} & \textbf{9.42 (-17\%)} & {5.65 (-1\%)} & \textbf{12.55 (-14\%)} \\ 
\bottomrule
\end{tabular}
}
\vspace{-5pt}\caption{Wikitext perplexity comparison on LLaMA-1, OpenLLaMA, and LLaMA-3.1 model families. * indicates results from either the previous paper \citep{sun2023simple}. - means results are not applicable from running their source code directly or out-of-memory issue. Bold highlights relative perplexity improvements over Wanda of 5\% or more.}
\label{tab:ppl}
\vspace{-5pt}
\end{table*}

\begin{figure}[h]
    \centering
    \includegraphics[width=0.49\textwidth]{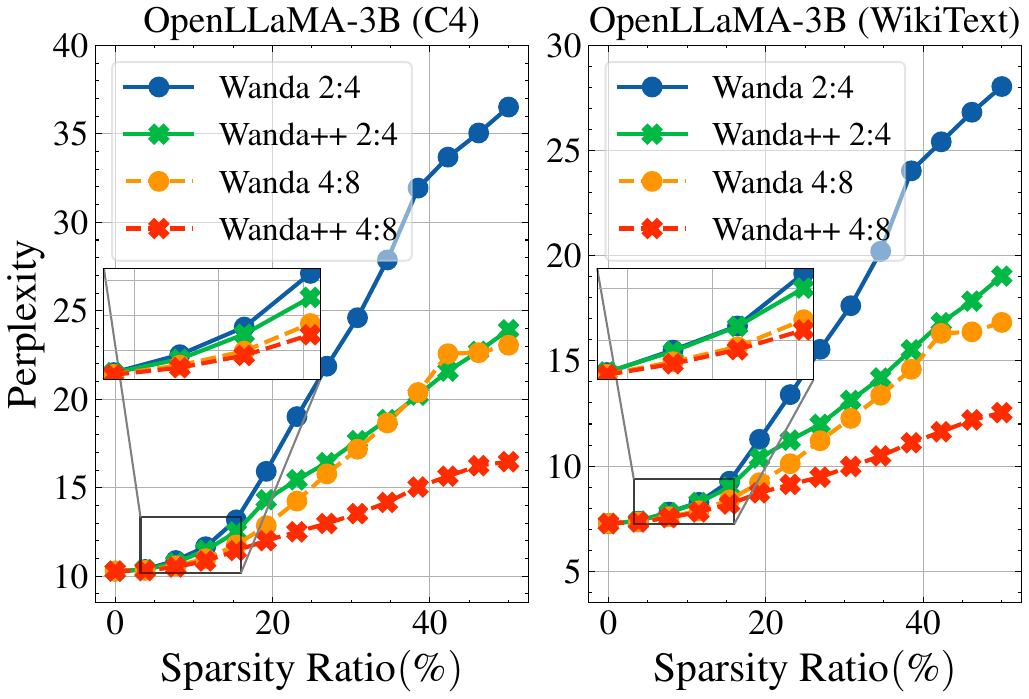}
    \vspace{-5pt}
    \caption{Perplexity on the C4 validation (left) and Wikitext test sets (right) as the sparsity ratio increases by gradually pruning two additional decoding blocks at a time, up to all 26 blocks. 
    Wanda++ leverages regional gradients and significantly outperforms Wanda: our 2:4 results are comparable to Wanda's 4:8 counterparts.}
    \label{fig:ro_curve}
    \vspace{-5pt}
\end{figure}

\subsection{Perplexity}\label{sec:lm}
First, we consider OpenLLaMA-3B, a relatively small model, for a more contrastive perplexity comparison between Wanda and our proposed method. As shown in Figure \ref{fig:ro_curve}, we begin by pruning the first 2 decoder blocks, then gradually apply N:M sparsity to two more decoder blocks until all 26 blocks are pruned. The LM head and embeddings are excluded from pruning. Perplexity results are reported on both the C4 validation dataset and the WikiText test dataset. When all decoder blocks are pruned with the N:M pattern, resulting in a 50\% sparsity ratio, our method outperforms Wanda by a noticeable margin: on the C4 validation dataset, with 2:4 sparsity, perplexity is reduced from 36.5 to 23.9, a relative reduction of 34.4\%; with 4:8 sparsity, perplexity is reduced from 23.1 to 16.4, a relative reduction of 29.0\%. Similarly, on the WikiText test dataset, the relative perplexity reductions are 32.1\% and 25.5\% for 2:4 and 4:8 sparsity, respectively. The margin generally increases with the sparsity ratio, although in the lower sparsity ratio region, particularly on WikiText (right plot), the benefit is obscure. Note that, in higher sparsity ratios, our method can achieve comparable or superior perplexity results under 2:4 sparsity compared to Wanda for 4:8 sparsity, a much more relaxed sparsity setting.

A more comprehensive comparison of perplexity is summarized in Table \ref{tab:ppl}, where LLaMA-1 models with four different sizes are included along with OpenLLaMA 7B and 70B. We consider SparseGPT and Wanda as two baseline pruning methods to compare with our method, Wanda++ w. RO. To understand the contribution of each enhancement in our method, we also report results from Wanda++ RO, where the Regional Optimizer (RO) is enabled for each decoder block, and Wanda++, where regional gradients are used along with the L2 norm of the input data for every layer in each decoder block. Regarding the sparsity patterns, we consider 50\% unstructured pruning, 2:4, and 4:8 semi-structured pruning methods, although 2:4 sparsity is the most commonly supported in hardware for runtime acceleration.

Between Wanda++ RO and Wanda++ RGS, leveraging regional gradients within the decoder block-level optimization proves more effective in mitigating pruning-induced degradation. Furthermore, applying RO to a better pruning criterion, as shown in Wanda++ w. RO compared to Wanda++ RO, further improves performance. We compare our method (the strongest among those three in italics) to Wanda and report the relative perplexity reductions in Table \ref{tab:ppl}. The most noticeable benefit is observed with 2:4 sparsity: on LLaMA-1, the relative reductions are 19\%, 20\%, 7\%, and 11\% for model sizes 7B, 13B, 30B, and 65B, respectively. Our method is most effective in reducing pruning degradation for smaller models, where higher pruning degradations usually occur, which is also the case for OpenLLaMA models. In all experiments, our method consistently shows superior perplexity compared to the baseline methods. However, for 4:8 sparsity and unstructured pruning, where perplexity values are generally lower than those with 2:4 sparsity, the benefits of our method become less salient. Nonetheless, the improvement remains more substantial compared to what Wanda achieves on top of SparseGPT.

Sparsity-aware model fine-tuning may further improve the performance. We further show our Wanda++ method is orthogonal to LoRA fine-tuning \cite{hu2021lora} method in Section. \ref{sec:lora}.  Also Another method distills smaller LLMs from a 15B pre-trained model but requires 16 DGX A100 nodes \citep{muralidharan2024compact} (128 80GB GPUs in total), whereas our method only needs one for the similar model size.

\begin{table*}[ht!]
\centering
\footnotesize
\setlength\tabcolsep{2.pt}
\resizebox{0.99\textwidth}{!}{%
\begin{tabular}{l|c|c|c|c|c|c|c|c|c}
\toprule
\textbf{Method} & \textbf{Wic} & \textbf{Mrpc} & \textbf{Hellaswag} & \textbf{Arc\_easy} & \textbf{Arc\_challenge} & \textbf{Winogrande} & \textbf{BoolQ} & \textbf{RTE} & \textbf{MMLU}  \\ \midrule
Baseline       & 49.84 & 69.12 & 56.96 & 75.29 & 41.80 & 70.00 & 75.02 & 66.43 & 35.10  \\ \midrule
Wanda          & 48.75 & 46.81 & 41.66 & 59.34 & 27.47 & 61.96 & 69.60 & 49.82 & 25.85  \\ \midrule
GBLM          & 49.32 & 65.31 & 41.80 & 61.43 & 30.45 & 63.24 & \textbf{71.20} & 57.43 & 26.34 \\ \midrule
\textit{Wanda++ RGS}        & 49.37 (1\%) & 64.46 (38\%) & 41.43 (-1\%) & 62.42 (5\%) & \textbf{31.06 (13\%)} & 62.83 (1\%) & 67.95 (-2\%) & 58.48 (17\%) & 26.40 (-2\%)  \\ \midrule
\textit{Wanda++ }  & \textbf{50.00 (2\%)} & \textbf{68.38 (46\%)} & \textbf{45.31 (8\%)} & \textbf{63.72 (7\%)} & {29.27 (6\%)} & \textbf{65.04 (4\%)} & {67.80 (-2\%)} & \textbf{62.09 (24\%)} & \textbf{27.52 (6\%)} \\ 
\bottomrule
\end{tabular}}
\vspace{-1pt}\caption{Zero-shot accuracy (\%) from LLaMA-1 7B across various tasks under 2:4 sparsity.}
\vspace{-2pt}
\label{tab:zero}
\vspace{-1pt}
\end{table*}
\begin{figure*}[t]
    \centering
    \includegraphics[width=0.95\textwidth]{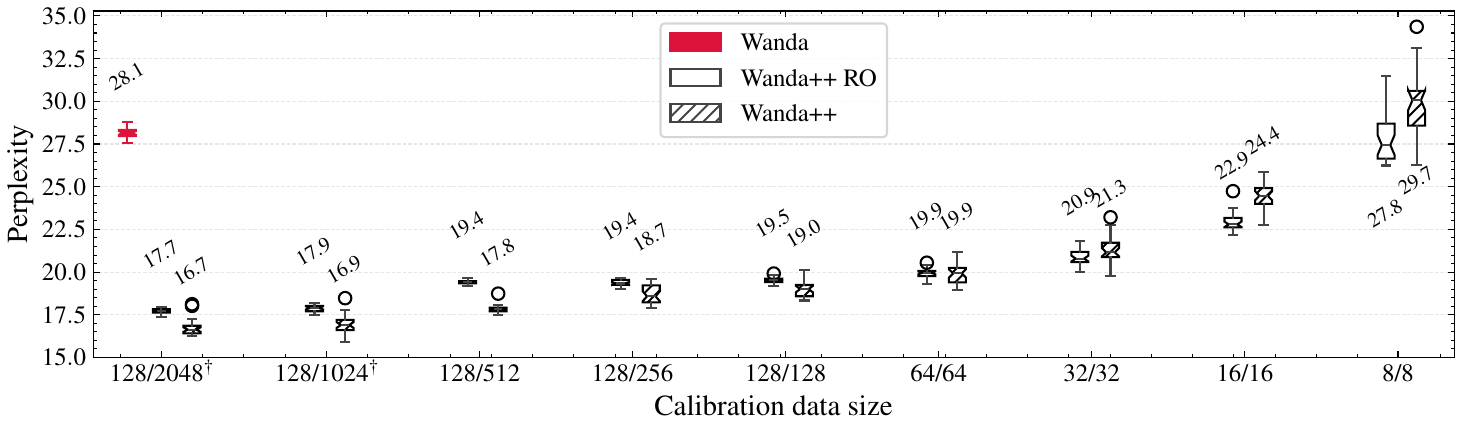}
    \vspace{-10pt}
    \caption{Box plot of perplexity on the Wikitext test set, based on 30 runs from 2:4 sparse OpenLLaMA-3B model.
    %larger calibration data generally improves performance of proposed methods. 
    }  
    \label{fig:token_efficiency}
\vspace{-15pt}
\end{figure*}
\subsection{Zero-Shot Accuracy}
We measure the zero-shot accuracy via Harness \cite{eval-harness}. Along with Wanda, we examine two of our proposed methods both leveraging regional gradients: \textit{Wanda++ RGS} only uses regional gradients to compute the pruning criterion, while \textit{Wanda++} applies them to both weight pruning (RGS) and in-block weight recompute (RO). We aim to understand if the observed benefit of improved perplexity from \textit{Wanda++} can pan out in downstream NLP tasks, and that the integration of RO does not negatively impact the generality of the pruned model. Results from the LLaMA 7B model on 9 tasks are listed in Table \ref{tab:zero}. All pruned models are conformed to 2:4 sparsity.
\\~\\
Although no consistent pattern shows, \textit{Wanda++} in general yields the best performance among the three pruning methods including Wanda which leads in \texttt{BoolQ} task. Compared to the margin from perplexity evaluation, the improvement from \textit{Wanda++} against \textit{Wanda++  RGS} is less salient. This is reasonable as RO is conducted on C4 dataset without optimizing any downstream tasks. Note that for \texttt{Mrpc} and \texttt{RTE} tasks, \textit{Wanda++} outperforms Wanda by 46\% and 24\%, respectively, close to the accuracy of the dense baseline.
\subsection{Sensitivity Analysis}
While Wanda’s complexity is $O(d_\text{hidden}^2)$, the pruning time and memory consumption both depend linearly on the amount of calibration data. This is also the case for our proposed methods. We alternate the number of samples and context length of each sample in C4 training data and compare the corresponding perplexities in each calibration dataset setting in Figure \ref{fig:token_efficiency} as the box plot. OpenLLaMA-3B is used in this sensitivity analysis on the size of calibration data.
We run each experiment 30 times. Each box extends from the lower to the upper quartile with a 95\% confidence interval (the notch) of the median. The outliers are also shown in the black circles. For Wanda, we stick to the default setting with 128 calibration samples and 2048 context length each. For both Wanda++ RO and Wanda++, we consider nine calibration settings (number of samples/context length): from a tiny calibration set of 8/8 up to the 128/2048 case. 
In both $128/2048^{\dagger}$ and $128/1024^{\dagger}$ settings, each epoch uses 32 random samples to avoid out-of-memory issues.
\\~\\
Compared to Wanda, which shows stable perplexity across various numbers of calibration samples \cite{sun2023simple}, our methods favor larger calibration sizes, particularly for Wanda++, which only starts to outperform Wanda++ RO beyond the 64/64 setting. However, even at the 16/16 setting, both of our proposed methods yield lower perplexities than Wanda. Both Wanda and Wanda++ RO are more stable overall than Wanda++. The comparison is less contrastive with larger calibration datasets.
%\\~\\
%In addition to the experimental results presented here, we conducted further experiments to evaluate the performance of the Wanda++ method in Appendix \ref{app:exp}, which include: 1) performance under higher sparsity 2) pruning time and memory usage comparison across different pruning methods 3) the latency and model size reduction achieved 4) an ablation study for hyperparameter selection in our RGS criterion.

\subsection{Pruning Time and Memory Consumption}\label{app:time}
We further discuss the memory and time efficiency of our proposed method. As mentioned earlier, integrating gradient information into the pruning score poses a significant computational challenge. Wanda avoids any gradient approximation and backpropagation, and therefore achieves simple and efficient LLM pruning compared with other post-training pruning methods like SparseGPT. Here, we evaluate the time and memory costs during the pruning process to demonstrate that our proposed method maintains similar computational advantages, especially when compared with previous LLMs pruning methods that utilize full model gradient information. For GBLM, we combined the time for both gradient computation and pruning, based on the provided code.

Without model weight updates, Wanda has the shortest pruning time. Wanda++ RGS (without RO) comes in second. When RO is added, as shown in the Wanda++ (M) row, the pruning time approaches that of SparseGPT. We also report metrics for Wanda++ (L) as a reference, though in practice, Wanda++ (M) is sufficient to achieve the performance shown in Table. \ref{tab:ppl}. It takes 10 minutes or less to prune the 7B and 13B models, and about 30 minutes for the 65B model. For 7B and 13B, one 80 GB GPU is enough, while the 65B model requires 4 H100 GPUs.

Unlike conventional BP involved methods that require loading the full model into memory, Wanda++ significantly reduces memory overhead by performing regional optimization at the decoder block level—loading only one decoder block at a time. This design decouples memory cost from the overall model size, which scales with both hidden dimension and the total number of decoder blocks. Instead, Wanda++’s memory usage depends solely on the hidden size. For instance, in the Megatron-Turing NLG 530B model \cite{smith2022using}—the largest model to our knowledge—each decoder block contains approximately 5.03 billion parameters with a hidden size of 20,480. The estimated theoretical memory required for optimizing a single block is approximately 40.24 GB (assuming we using Adam \cite{kingma2014adam} optimizer). While actual usage may be higher due to activations and implementation overhead, it remains within twice this estimate and can be accommodated by a single 80GB GPU with proper optimization. This modest memory cost is particularly favorable given the up to 32\% performance improvement achieved by Wanda++, especially when contrasted with the 16–20 80GB GPUs typically needed just for inference of such large-scale models.
%As indicated in Table \ref{tab:eff}, the pruning time for Wanda++ is 40 times lower than that of the GBLM method for a 2:4 sparsity case across OpenLLaMA-7B/13B/70B models, while keeping the memory cost the same. 
%Additionally, Wanda++ achieves a speedup of 5 times compared with the SparseGPT method. 
%It is reasonable that the proposed method is slightly slower than the Wanda method. However, we believe the difference of 5 to 10 minutes in pruning time between Wanda and Wanda++ on a 70B model does not significantly impact the choice of methods, given the remarkable improvement of Wanda++ in pruning performance.
\begin{table}[h]
\centering
\footnotesize
\setlength\tabcolsep{4.pt}
\begin{tabular}{c|ccc|ccc}
\toprule
\multirow{2}{*}{\textbf{Method}} & \multicolumn{3}{c|}{\textbf{Time (Sec.)}} & \multicolumn{3}{c}{\textbf{Memory (GB)}} \\ %\cline{2-7} 
\cmidrule(lr){2-7}
                         & 7B         & 13B        & 65B      & 7B       & 13B      & 65B       \\ \midrule
SparseGPT                & 322        & 594      &     -     & 23       & 38       &    -       \\
GBLM                     & 5801       & 10733      &     49663     & 26       & 50       &   269        \\
Wanda                    & 55         & 95         &   628    & 22       & 36       &    320    \\
\textit{Wanda++ RGS}                  & 147        & 190        &  1461     &    29      &     49     &    320       \\ 
\textit{Wanda++ (M)}        &  290       &   574      &  1821     &   25       &    49      &    280       \\
\textit{Wanda++ (L)}        &     2381    &    5569     &    22409   &    31      &    49      &335           \\
\bottomrule
\end{tabular}
\caption{Memory and pruning time comparison: For Wanda++, we consider two calibration settings that differ in the nunmber of tokens per input sample. \textit{Wanda++ (M)} uses an input length of 128, while  \textit{Wanda++ (L)} uses 2048 like others.}
\vspace{-15pt}
\label{tab:eff}
\end{table}

\subsection{Sparsity-aware Fine-tuning}\label{sec:lora}
We conducted experiments using LoRA \cite{hu2021lora} to fine-tune both Wanda and Wanda++ pruned LLaMA-1 7B models. We followed the same experimental settings as the original Wanda paper, where LoRA is applied to the q and v modules in all transformer blocks. Both models were trained for 30k steps on the C4 dataset. As shown in Table \ref{tab:lora}, our method achieves similar improvements with LoRA compared to the Wanda method. This demonstrates our proposed method is orthogonal to the fine-tuning approach, further strengthening the fairness of our comparison with weight-update-free methods like Wanda in Table \ref{tab:ppl}. Note that even we fine-tune the model with LoRA, the process is still time consuming, which takes around 12 hours to improve the performance. Also, the LoRA fine-tuning method requires much more memory compared with regional optimization, which makes it infeasible for the weights update of large scale pruned models.
\begin{table}
% \vspace{-2mm}
\centering
\footnotesize
\setlength\tabcolsep{4.pt}
\begin{tabular}{c|ccc}
\toprule
Methods & Dense & 	Pruned Model & 	 After LoRA-tuned \\  \midrule
Wanda       & 5.68   & 11.59  & 8.23(-29\%)  \\ 
Wanda++     & 5.68   & 9.43   & 6.88(-27\%)     \\
\bottomrule
\end{tabular}
\caption{Perplexity comparison on Wikitext with LoRA. All experiments are conducted on LLaMA-7B model with 2:4 sparsity.}
\vspace{-15pt}
\label{tab:lora}
\end{table}

\subsection{Experiments on Higher Sparsity}
In this section, we evaluate the performance of our proposed method under higher unstructured sparsity levels. Specifically, we compare Wanda++ against baselines such as GBLM and the original Wanda, with results summarized in Table~\ref{tab:higher}. As shown, Wanda++ consistently outperforms prior methods across varying sparsity levels. However, we note that unstructured pruning—even at high sparsity—may not yield inference speedups comparable to structured formats such as 2:4 sparsity on modern GPUs. As a result, the 2:4 and 4:8 sparsity settings are of greater practical relevance for deployment.
\begin{table}[h]
    \centering
    \footnotesize
    \begin{tabular}{lccc}
        \hline
        Methods & 0.6 & 0.7 & 0.8 \\
        \hline
        GBLM & 10.37 & 54.60 & 2550.10 \\
        Wanda & 9.71 & 76.17 & 1942.53 \\
        Wanda++ & 9.50 & 55.52 & 1586.69 \\
        \hline
    \end{tabular}
    \caption{Performance comparison of different methods under varying sparsity ratios.}
    \vspace{-15pt}
    \label{tab:higher}
\end{table}

\section{Extending Wanda++ to Structured Pruning}
While our main focus is on unstructured and semi-structured pruning, we further evaluate the applicability of our method to structured pruning to inspire the future work. Specifically, we conduct experiments on a naive row-wise structured pruning (SP) for each weight matrices in the model, where the pruning score for each row is the average score of all paramters in the row. Prior work has shown that naively adapt Wanda methods to structured pruning (Wanda-SP) tend to degrade sharply under structured settings when applied naively \cite{an2024fluctuation}. Nonetheless, we adapt our approach (Wanda++-SP) to this setting and compare it against a baseline Wanda-SP method across varying pruning ratios.
\begin{table}[h]
\centering
\footnotesize
\begin{tabular}{cccc}
\toprule
Method & 0.1 & 0.3 & 0.5 \\
\midrule
Wanda-SP & 9.10 & 118.72 & 8804.84 \\
Wanda++-SP & 7.84 & 63.45 & 5428.97 \\
\bottomrule
\end{tabular}
\caption{Perplexity of row-wise structured pruning (SP) on LLMs at different sparsity levels. Lower is better.}
\vspace{-15pt}
\label{tab:structured}
\end{table}

As shown in Table \ref{tab:structured}, Wanda++-SP consistently achieves lower perplexity than Wanda-SP, particularly at higher pruning ratios. These results highlight the robustness and generalizability of our method beyond the unstructured and semi-structured regimes originally explored in this work. We further emphasize that the regional optimization component of Wanda++ is orthogonal to most existing structured pruning methods, such as SliceGPT~\cite{ashkboos2024slicegpt} and LLM-Pruner~\cite{ma2023llm}. In practice, a lightweight regional optimization using only a few hundred calibration samples is sufficient to yield substantial improvements in model performance.
\section{Conclusion}
%In this paper, we achieve remarkable improvement for LLMs pruning by considering the regional gradient calculated within each decoder. We propose a new pruning framework named Wanda++, which includes a novel regional gradient score to decide the importance of each weight parameter and a regional optimization method to slightly update the pruned weights. Experimental results show that our proposed pruning framework significantly reduced the pruning degradation by considering the regional information.
In this paper, we proposed Wanda++, a lightweight post-training LLM pruning method that leverages regional gradients to effectively mitigate pruning-induced performance degradation. Wanda++ defines a region as each decoder block. The method includes a pruner based on regional gradient scores (RGS) and a Regional Optimizer (RO) for in-block, sparsity-aware calibration. By utilizing regional gradients, it outperforms Wanda on various LLaMA models. Wanda++ is efficient, especially compared to full gradient fine-tuning methods, pruning 7B LLMs in 10 minutes, as it operates only within each decoder block.

We also note that the regional optimization component of Wanda++ is \textbf{orthogonal} to most existing LLM pruning approaches, as well as to works exploring dense-to-MoE transformations \cite{wu2024parameter}. Incorporating regional optimization from Wanda++ provides a strong initialization for the foundation model, which can be further combined with downstream optimization techniques such as LoRA fine-tuning \cite{hu2021lora}, Direct Preference Optimization (DPO) \cite{rafailov2023direct}, or Group Relative Policy Optimization (GRPO) \cite{rafailov2023direct}. An exciting direction for future work is to explore the potential of using Wanda++ as a general tool to enhance model performance following architectural adaptations.
\section*{Limitations}
While Wanda++ demonstrates significant improvements in mitigating pruning-induced degradation and achieves state-of-the-art results under unstructured, 2:4, and 4:8 sparsity patterns, we have not fully tested Wanda++ for structured pruning, as it is not the main focus of this paper. Although the RO method has been tested with structured pruning approaches like SliceGPT, the full Wanda++ framework is not yet fully compatible with structured pruning. Future work could focus on adapting the Wanda++ method to better support structured pruning scenarios.
\section*{Acknowledgment}
We thank Denis Filimonov for his insights in developing the initial idea and refining the RGS score.
 \vspace{-5pt}

\bibliography{custom}

\appendix
\begin{algorithm*}[h]
\caption{PyTorch pseudo-code for computing RGS criterion for a single decoder block}
\label{alg:pseudo}
\vspace{-5pt}
\begin{lstlisting}[style=python, numbers=none]
# block: DecoderBlock
# inps: Input hidden states of decoder block
# sq_grad: empty gradient dictionary
# alpha: hyperparameter
def wanda_plus_plus_pruner(block: torch.nn.module, inps: List, sq_grad: Dict, alpha: Constant):
    for inp in inps:
        loss = torch.norm(block.forward(inp))
        loss.backward()
        with torch.no_grad():   # Compute aggregate gradient for each weight
            for name, param in block.named_parameters():
                sq_grad[name] += param.grad ** 2
    for key, value in sq_grad.items():
        sq_grad[key] = torch.sqrt(value / inps_len)
    for n, par in layer.named_parameters(): # Compute the pruning scores
        layer_inps = torch.stack(get_signal(inps, n)) # Obtaining and stacking layer-wise inputs
        score = par.weight.abs() * (alpha * sq_grad[n] +  layer_inps.norm(p=2, dim=0))
\end{lstlisting}
\vspace{-10pt}
\end{algorithm*}
\section{Pytorch pseudo-code for Calculating RGS Criterion}\label{app:algotihms}
For the pruning process,  we first perform the forward and construct the loss function with the L2 norm of the decoder output hidden state. Then, a backward pass is conducted to obtain the gradient for each parameter weight matrices and the scaled stochastic gradients regarding each calibration data sample are aggregated to reduce the noise included with each single data sample. The aggregated gradient magnitude is normalized with the number of calibration data samples and used to calculate the designed pruning score with the weight magnitude given, as given in Eq. (\ref{eq:rgs}). Also, we summarize the Pytorch pseudo-code for calculating our RGS criterion here.

\section{Additional Experimental Results}\label{app:exp}
%\subsection{Higher Sparsity}
%In addition to evaluating our proposed method on standard unstructured pruning with a 0.5 sparsity level, we also test its robustness at higher sparsity ratios. The results are reported in Table \ref{table:high} for language modeling tasks based on the experimental setup introduced in Sec. \ref{sec:lm} using LLaMA-1-7B models Based on these results, we conclude that our Wanda++ method still achieves remarkable improvements even in extreme cases compared to previous post-training pruning methods like Wanda. This expands the potential use cases for our proposed method.
%\begin{table}[h]
%\footnotesize
%\setlength\tabcolsep{6.pt}
%\centering
%\caption{Evaluation on higher sparsity}
%\begin{tabular}{c|ccc}
%\toprule
%\multicolumn{1}{l|}{\textbf{Method}\textbackslash{}\textbf{Sparsity}} & 0.6   & 0.7   & 0.8     \\ \midrule
%SparseGPT                                     &    10.39       &  25.85     &    195.74     \\
%GBLM                                              &       &       &         \\
%Wanda                                             & 14.99 & 85.57 & 2449.34 \\
%Wanda++                                           & 9.70   & 55.52 & 1586.69 \\ \bottomrule
%\end{tabular}
%\label{table:high}
%\end{table}

\subsection{Model Size and Latency Reduction}
%While it is known that 2:4 sparsity is supported via Sparse Tensor Cores on NVIDIA GPUs \cite{nvidia_sparsity_int8}, 
We measure the Time to First Token (TTFT), Time Per Output Token (TPOT) and total model weight memory consumption to examine 2:4 sparsity's actual impact on a dummy 7B LLaMA-akin model in Table \ref{tab:reduction_metrics} using TensorRT-LLM-0.9.0 with the Sparse Tensor Core support \cite{nvidia_sparsity_int8}. Only the multi-layer perceptron (MLP) modules are pruned, with both tensor parallelism and pipeline parallelism set to 1. Under FP16 format, we observe a TTFT reduction of 33\% or more, while the TPOT reduction is around 10\%. Total weight memory is reduced by 28\% for FP16 (from 12.8 GB to 9.2 GB), which are also reflected in the size of compiled TensorRT engines. See Appendix \ref{app:fp8} for FP8 format \cite{kuzmin2022fp8} results.

%For FP8, the TTFT latency reduction is smaller, particularly when the batch size and input length increase. One explanation is that the model leans towards being compute-bound, where reducing weight memory load becomes less meaningful. TPOT reduction under FP8 is 13\% or greater, except when the batch size is 4 and the output length is 4096. 
%Total weight memory is reduced by 28\% for FP16 (from 12.8 GB to 9.2 GB) and 22\% for FP8 (from 6.8 GB to 5.3 GB), which are also reflected in the size of compiled TensorRT engines.

\begin{table}[t]
\centering
\footnotesize
\setlength\tabcolsep{7pt}
\resizebox{0.48\textwidth}{!}{
\begin{tabular}{c|c|c|c|c|c}
\toprule
\multirow{2}{*}{\makecell{\textbf{Batch} \\ \textbf{Size}}} 
& \multicolumn{2}{c|}{\textbf{Token Length}} %\multirow{2}{*}{\makecell{\textbf{Input} \\ \textbf{Length}}}
 %& \multirow{2}{*}{\makecell{\textbf{Output} \\ \textbf{Length}}}  
& \multicolumn{2}{c|}{\textbf{Latency}} 
  %&\multirow{2}{*}{\makecell{\textbf{TPOT} \\ \textbf{Latency}}}  
  &\multirow{2}{*}{\makecell{\textbf{Weight} \\ \textbf{Memory}}}  
  \\
  \addlinespace[1pt]\cline{2-5}\addlinespace[1pt]
  %\\
  &\textbf{Input}&\textbf{Output}&\textbf{TTFT}&\textbf{TPOT}& \\
  \midrule
\multirow{6}{*}{1} & 128  & \multirow{6}{*}{64}  & 33 & 10 & \multirow{11}{*}{\centering 28} \\  \cmidrule(lr){2-2} \cmidrule(lr){4-5}
                   & 1024 &  & 47 & 11 &  \\ \cmidrule(lr){2-2} \cmidrule(lr){4-5}
                   & 2048 &  & 47 & 10 &  \\\cmidrule(lr){2-2} \cmidrule(lr){4-5}
                   & 4096 &  & 46 & 10 &  \\ \cmidrule(lr){1-5} \cmidrule(lr){4-5}
\multirow{6}{*}{4} & 128  & \multirow{6}{*}{64}  & 45 & 11 &  \\ \cmidrule(lr){2-2} \cmidrule(lr){4-5}
                   & 1024 &  & 47 & 11 &  \\ \cmidrule(lr){2-2} \cmidrule(lr){4-5}
                   & 2048 &  & 45 & 9  &  \\ \cmidrule(lr){2-2} \cmidrule(lr){4-5}
                   & 4096 &  & 43 & 7  &  \\ \bottomrule
\end{tabular}
}
\vspace{-1pt}
\caption{Relative reduction (\%) for latency and weight memory from 2:4 sparsity under FP16 format.}
\vspace{-15pt}
\label{tab:reduction_metrics}
\end{table}

\subsection{Ablation Study on RGS Scaling Factor}\label{app:alpha}
%Although in \citep{das2023beyond}, 
We examine the hyperparameter $\alpha$ in the RGS criterion (Eq. \ref{eq:rgs}), which balances the regional gradient score and the layer-wise Wanda score. An ablation study is conducted, testing $\alpha$ values from 1 to 1,000,000, to assess their effect on perplexity. Results from LLaMA-3 8B models are presented in Table \ref{tab:alpha}. The lowest perplexity for LLaMA-3 8B with 2:4 sparsity occurs at $\alpha=50$, indicating that the optimal choice of $\alpha$ is model-specific.
\begin{table}[h]
\centering
\footnotesize
\setlength\tabcolsep{7pt}
\begin{tabular}{lc}
\toprule
RGS Criterion & Perplexity \\
\midrule
  $(1\cdot \bm{G}_{ij} + \|\bm{X}_j\|_2)\cdot |\bm{W}_{ij}|$           &                 24.55                    \\
$(10\cdot \bm{G}_{ij} + \|\bm{X}_j\|_2)\cdot |\bm{W}_{ij}|$       &              24.62                        \\
$(50\cdot \bm{G}_{ij} + \|\bm{X}_j\|_2)\cdot |\bm{W}_{ij}|$      &               23.99                    \\
$(100\cdot \bm{G}_{ij} + \|\bm{X}_j\|_2)\cdot |\bm{W}_{ij}|$      &             24.68                \\
$(500\cdot \bm{G}_{ij} + \|\bm{X}_j\|_2)\cdot |\bm{W}_{ij}|$      &            25.66                \\
$(1000\cdot \bm{G}_{ij} + \|\bm{X}_j\|_2)\cdot |\bm{W}_{ij}|$      &           26.25                \\
$(5000\cdot \bm{G}_{ij} + \|\bm{X}_j\|_2)\cdot |\bm{W}_{ij}|$      &           29.07                \\
$(10000\cdot \bm{G}_{ij} + \|\bm{X}_j\|_2)\cdot |\bm{W}_{ij}|$      &          29.90                \\
$(1000000\cdot \bm{G}_{ij} + \|\bm{X}_j\|_2)\cdot |\bm{W}_{ij}|$      &        31.14                    \\
         \bottomrule
\end{tabular}
\caption{Perplexity with different $\alpha$ values in RGS on LLaMA-3 8B model for 2:4 sparsity.}
\vspace{-5pt}
\label{tab:alpha}
\end{table}
\subsection{Latency / Model Size Reduction for FP8}\label{app:fp8}
When the weight, activation and KV-cache are quantized to the FP8 format, the TTFT latency reduction from 2:4 sparsity is smaller, particularly when the batch size and input length increase compared to that under FP16. One explanation is that the model leans towards being compute-bound, where reducing weight memory load becomes less meaningful. TPOT reduction under FP8 is 13\% or greater, except when the batch size is 4 and the output length is 4096. 
Total weight memory is reduced by 22\% with 2:4 sparsity under the FP8 format (from 6.8 GB to 5.3 GB).
\begin{table}[h]
\centering
\footnotesize
\setlength\tabcolsep{7pt}
\resizebox{0.48\textwidth}{!}{
\begin{tabular}{c|c|c|c|c|c}
\toprule
\multirow{2}{*}{\makecell{\textbf{Batch} \\ \textbf{Size}}} 
& \multicolumn{2}{c|}{\textbf{Token Length}} %\multirow{2}{*}{\makecell{\textbf{Input} \\ \textbf{Length}}}
 %& \multirow{2}{*}{\makecell{\textbf{Output} \\ \textbf{Length}}}  
& \multicolumn{2}{c|}{\textbf{Latency}} 
  %&\multirow{2}{*}{\makecell{\textbf{TPOT} \\ \textbf{Latency}}}  
  &\multirow{2}{*}{\makecell{\textbf{Weight} \\ \textbf{Memory}}}  
  \\
  \addlinespace[1pt]\cline{2-5}\addlinespace[1pt]
  %\\
  &\textbf{Input}&\textbf{Output}&\textbf{TTFT}&\textbf{TPOT}& \\
  \midrule
\multirow{6}{*}{1} & 128  & \multirow{6}{*}{64}  & 15 & 15 & \multirow{11}{*}{\centering 22} \\  \cmidrule(lr){2-2} \cmidrule(lr){4-5}
                   & 1024 &  & 4  & 13 &  \\ \cmidrule(lr){2-2} \cmidrule(lr){4-5}
                   & 2048 &  & 7  & 15 &  \\\cmidrule(lr){2-2} \cmidrule(lr){4-5}
                   & 4096 &  & 8  & 14 &  \\ \cmidrule(lr){1-5} \cmidrule(lr){4-5}
\multirow{6}{*}{4} & 128  & \multirow{6}{*}{64}  & 7  & 16 &  \\ \cmidrule(lr){2-2} \cmidrule(lr){4-5}
                   & 1024 &  & 7  & 14 &  \\ \cmidrule(lr){2-2} \cmidrule(lr){4-5}
                   & 2048 &  & 0  & 13 &  \\ \cmidrule(lr){2-2} \cmidrule(lr){4-5}
                   & 4096 &  & -13 & 1 &  \\ \bottomrule
\end{tabular}
}
\vspace{-5pt}
\caption{Relative reduction (\%) for latency and weight memory from 2:4 sparsity under FP8 format.}
\vspace{-15pt}
\label{tab:reduction_metrics_fp8}
\end{table}

\end{document}